\documentclass[fleqn,10pt]{wlscirep}
\usepackage[utf8]{inputenc}
\usepackage[justification=justified]{caption} 
\usepackage[T1]{fontenc}

\renewcommand{\mathcal}[1]{\mathrm{#1}}
\usepackage{amsthm}   
\theoremstyle{plain}

\usepackage{lineno} 

\theoremstyle{remark}

\usepackage{float}  
\title{Logic-informed reinforcement learning for cross-domain optimization of large-scale cyber-physical systems}

\author[1]{Guangxi Wan}
\author[1,*]{Peng Zeng}
\author[1,2]{Xiaoting Dong}
\author[1]{Chunhe Song}
\author[1]{Shijie Cui}
\author[1]{Dong Li}
\author[1]{Qingwei Dong}
\author[1]{Yiyang Liu}
\author[1]{Hongfei Bai}

\affil[1]{State Key Laboratory of Robotics and Intelligent Systems, Shenyang Institute of Automation, Chinese Academy of Sciences, Shenyang 110016, China}
\affil[2]{University of Chinese Academy of Sciences, Beijing 100049, China}

\affil[*]{Corresponding: zp@sia.cn}

\begin{abstract}
Cyber–physical systems (CPS) require the joint optimization of discrete cyber actions and continuous physical parameters under stringent safety logic constraints.  However, existing hierarchical approaches often compromise global optimality, whereas reinforcement learning (RL) in hybrid action spaces often relies on brittle reward penalties, masking, or shielding and struggles to guarantee  constraint satisfaction. We present logic-informed reinforcement learning (LIRL), which equips standard policy-gradient algorithms with projection that maps a low-dimensional latent action onto the admissible hybrid manifold defined on-the-fly by first-order logic. This guarantees feasibility of every exploratory step without penalty tuning. Experimental evaluations have been conducted across multiple scenarios, including industrial manufacturing, electric vehicle charging stations, and traffic signal control, in all of which the proposed method outperforms existing hierarchical optimization approaches. Taking a robotic reducer assembly system in industrial manufacturing as an example, LIRL achieves a 36.47\%–44.33\% reduction at most in the combined makespan–energy objective compared to conventional industrial hierarchical scheduling methods. Meanwhile, it consistently maintains zero constraint violations and significantly surpasses state-of-the-art hybrid-action reinforcement learning baselines. Thanks to its declarative logic-based constraint formulation, the framework can be seamlessly transferred to other domains such as smart transportation  and smart grid, thereby paving the way for safe and real-time optimization in large-scale CPS.

\end{abstract}
\begin{document}

\flushbottom
\maketitle
%
%
\thispagestyle{empty}


Cyber-physical systems (CPS) seamlessly integrate sensing, computation, and actuation to enable closed-loop control of large-scale physical processes\cite{A1,A2,A3,A4}. Prominent examples, such as smart factories\cite{A6}, autonomous transportation networks\cite{A5}, and wide-area power grids\cite{A7}, constitute crucial components of modern industry and infrastructure. In these applications, the overall performance of the system is critically based on cross-domain optimization\cite{A8}. At each decision point, the supervisory controller must simultaneously (i) schedule discrete cyber issues, including task assignments, mode switches, and component start-ups, and (ii) adjust continuous physical parameters, such as robot trajectories, vehicle accelerations, or generator set-points. These two aspects are interlinked by strict safety, logical, and resource constraints. Violations can result in severe system-wide failures. Therefore, attaining real-time and globally optimal decision-making within such hybrid decision spaces poses a fundamental challenge in both science and engineering. 

\begin{figure}[!ht]
	\centering
	\includegraphics[width=\linewidth]{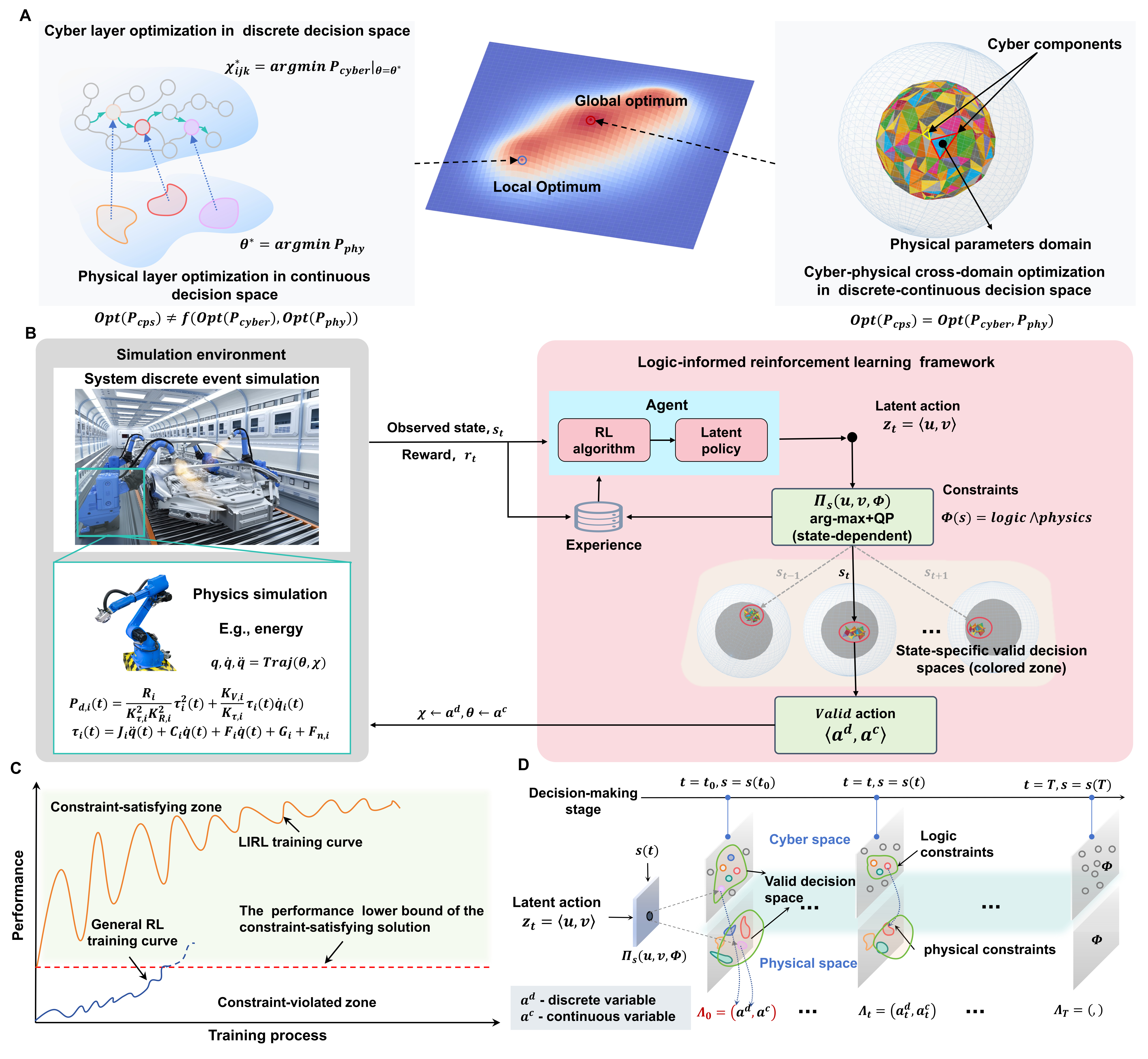}
	\caption{Overview of CPS cross-domain optimization and LIRL framework. \textbf{A}. The concept of Hierarchical optimization and cross-domain optimization. Hierarchical optimization struggles to achieve the global optimum due to the lack of a strict optimal substructure between levels, while cross-domain optimization can attain the globally optimal solution. \textbf{B}. The LIRL introduces two key innovations: latent action projection and dynamic valid-action-space partitioning. It first maps hybrid discrete-continuous actions into a unified latent continuous space, then projects them via function $\Pi_s$ into the valid explicit action space—which is adaptively partitioned based on system state and constraints. \textbf{C}. The characteristics of the LIRL training process guarantee the continuous satisfaction of constraints throughout the training period, while steadily enhancing performance.  \textbf{D}. The process of latent action projection and valid action space partitioning, The valid action space represents the constrained solution space under varying CPS system states.  Decision-making activates only when the valid action space is non-empty and halts automatically otherwise. }
	\label{fig1}
\end{figure}

In contemporary industrial practices, a significant reliance is placed on hierarchical decoupling\cite{A9}.  Specifically, a high-level planner first devises a discrete schedule and then conveys it to a low-level optimizer, or the process may occur in the reverse order. This low-level optimizer refines continuous parameters under the premise of fixed logic. Although this approach is computationally manageable, it unavoidably forfeits global optimality, as shown in Figure \ref{fig1}A . This is primarily because the optimization of each layer is often based on idealized assumptions, failing to adequately account for mutual coupling constraints and dynamic interactions. Moreover, no optimal substructure exists between the cyber and physical layer optimizations in the decomposed sub-problems.  As manufacturing, traffic, and energy systems continue to expand on a scale, the performance degradation that stems from decoupling becomes increasingly prominent\cite{A10}. 

Reinforcement learning (RL) presents an appealing alternative as, in theory, it can learn direct mappings from system state to hybrid actions without explicitly modeling the underlying dynamics. RL has achieved superhuman performance in various domains, including games, robotic manipulation, and the energy management of single electric vehicles. However, its application to large-scale CPS has been hampered by two persistent challenges. Firstly, the decision space in such scenarios is hybrid and high-dimensional. The discrete and continuous components demand distinct policy representations, and the size of their Cartesian product expands exponentially with the growth of the system. Specialized architectures like parameterized deep Q-Network\cite{A22} (PDQN), hierarchical proximal policy optimization\cite{A23} (HPPO), or multi-branch actor-critic networks can address specific subsets of hybrid problems. However, these architectures suffer from limited generality and complex network design requirements. Secondly, RL depends on reward signals to drive exploration. In CPS, objectives, such as minimizing a weighted combination of energy consumption, throughput, and safety risks, are intricately linked with strict logical and physical constraints. Encoding these constraints into a single scalar reward requires a delicate balance between positive goals and negative penalties. Incorrect specification often results in either continuous constraint violations or overly cautious policies that sacrifice performance. Although augmenting RL with action masking or shielding can alleviate immediate violations, it introduces discontinuous gradients, which impede convergence and may lead to short-sighted corrective actions. 

Recent investigations in the realm of logic-guided or safe RL have incorporated temporal-logic templates or safety layers into policy learning, thus ensuring the satisfaction of constraints\cite{A11,A12}. Nevertheless, the majority of these approaches are tailored either to purely discrete actions\cite{A13,A14} or to purely continuous control with differentiable constraints\cite{A15,A16,A17}. They do not offer a unified approach for hybrid CPS and do not analyze how the imposed corrections interact with gradient-based policy updates. 

In light of these gaps, we propose logic-informed reinforcement learning (LIRL), a general framework shown in Figure \ref{fig1}B that enhances off-the-shelf policy-gradient algorithms to optimize large-scale CPS under strict adherence to hybrid constraints. The core idea is to decouple exploration from feasibility through a latent unified action space. At each decision-making step, the agent proposes a low-dimensional latent vector, which is then projected onto the admissible manifold defined by first-order logic incorporating both cyber and physical constraints. This projection operation (i) guarantees that all executed actions are feasible, (ii) preserves gradient smoothness in continuous sub-spaces, and (iii) eliminates the need for reward shaping or pre-trained autoencoders. This paradigm ensures continuous satisfaction of constraints throughout training, while steadily improving performance, as shown in Figure \ref{fig1}C. Unlike traditional RL methods, where exploratory actions often cause early constraint violations, our approach guarantees strict constraint compliance from the very beginning, even when using random policies. As Figure  \ref{fig1}D illustrates, we define a state-specific valid action space. The presence of valid actions determines the decision point: optimization is triggered when available; otherwise, the cycle terminates. This mechanism restricts the agent to only generate valid actions, thereby concentrating exploration entirely on feasible policies. By updating the policy solely with valid actions, the policy gradient is guided toward high-reward and constraint-satisfying regions, accelerating convergence. Since the logical specification is declarative, the same reinforcement learning backbone can be efficiently applied to various CPS domains with minimal engineering effort. 

Our approach delivers four main contributions in this paper: (i)The cross-domain optimization of CPS is formulated as a constrained Markov decision process\cite{A18} with hybrid actions, and a projection-based method is presented. This method transforms the latent action into a hybrid action that satisfies the constraints in a closed form. (ii) Under mild Lipschitz conditions, the projection operator is proven to maintain the fixed points of standard policy-gradient updates, thereby preserving the convergence guarantees of the underlying RL algorithm. (iii) LIRL was implemented on a robotic reducer-assembly manufacturing system and evaluated in four optimization scenarios designed to jointly minimize makespan and energy consumption. Compared with state-of-the-art hierarchical schedulers, LIRL achieves an improvement of 36.47\%–44.33\% at most in aggregate reward. It consistently outperforms recent hybrid-action RL baselines and completely avoids constraint violations throughout the entire training process. (iv) Extensive ablation studies and cross-domain case studies—including elevator door header assembly, traffic signal coordination, and electric vehicle (EV) charging station scheduling—were systematically conducted to validate the generality, sample efficiency, and scalability of LIRL.

\section*{Results}

\begin{figure}[ht]
\centering
\includegraphics[width=\linewidth]{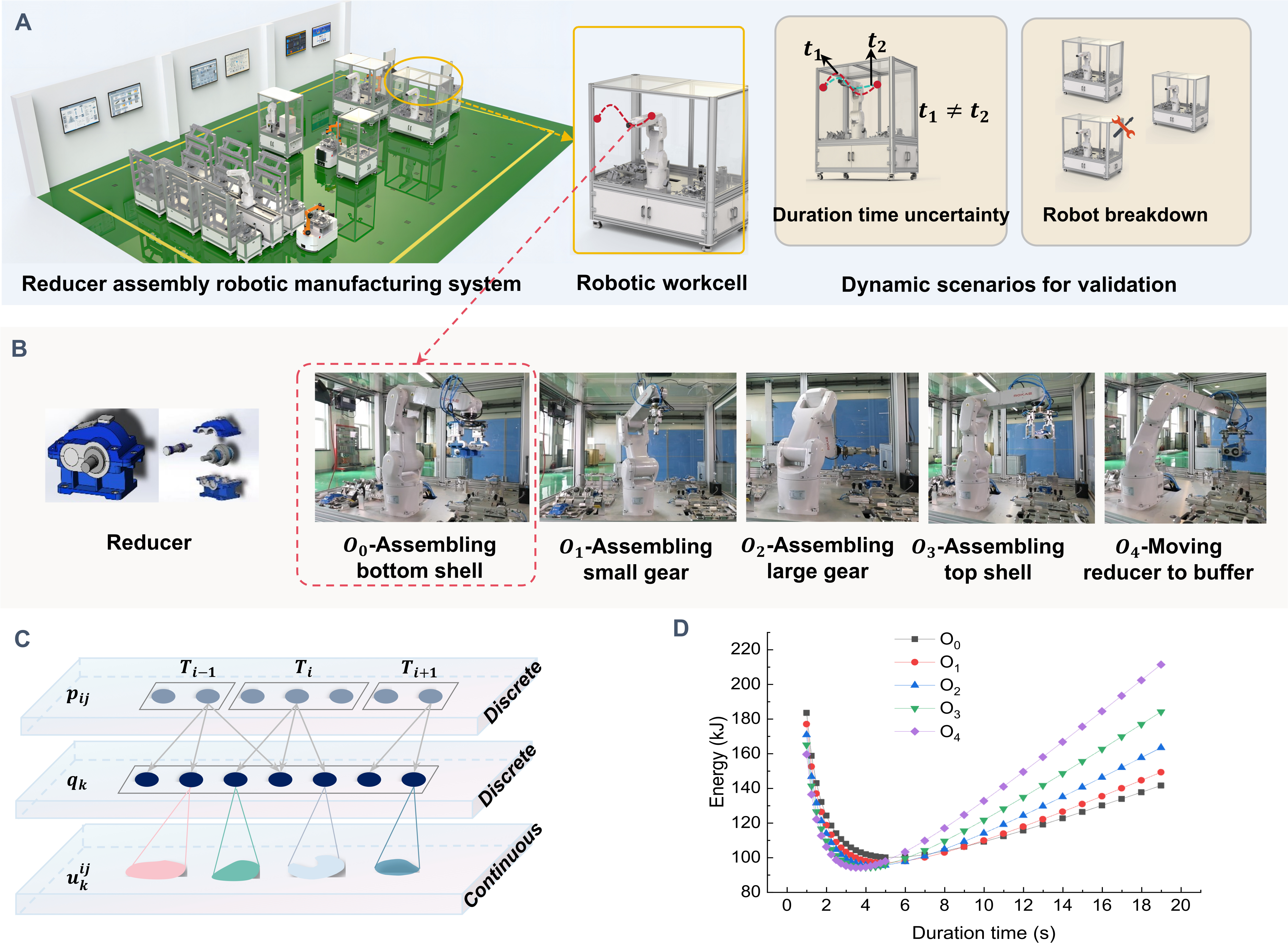}
\caption{Overview of $R^2AMS$.  \textbf{A}. $R^2AMS$ consists of modular robotic workcells, each considered a subsystem. To evaluate the generalization capability of LIRL, we designed two dynamic scenarios: task duration uncertainty and robot breakdown. \textbf{B}. Five steps for reducer assembly:  \textit{place bottom shell}, \textit{small gear assembling}, \textit{large gear assembling}, \textit{top bottom shell} and \textit{move reducer to buffer}. \textbf{C}. The structure of hybrid decision-making space. $T_i$ is the $i\mbox{-}th$ reducer; $p_{ij}$ is the $j\mbox{-}th$ stage of the $i\mbox{-}th$ reducer assembly task; $q_k$ is the $k\mbox{-}th$ robot workcell; $u_k^{ij}$ represents the trajectory configuration parameters of the robot workcell $q_k$  when completing $p_{ij}$. \textbf{D}. The energy signature of a robot workcell to process five operations of reducer assembling. The horizontal axis corresponds to the action processing time, where shorter times correspond to faster robot execution.}
\label{fig2}
\end{figure}

Empirical evidence from the simulation of a robotic reducer-assembly manufacturing System ($R^2AMS$) (Figure \ref{fig2}) demonstrates the effectiveness, efficiency, and robustness of the LIRL approach. All code, configuration files, and raw logs are publicly accessible at \url{https://github.com/wanguangxi/LIRL-CPS}. Details of additional experiments carried out in a real elevator door assembly industry, a simulated transportation scenario, and a simulated micro-grid scenario are provided in the Supplementary Material.

\subsection*{Overview of the robotic reducer-assembly system}

Gear assembly accounts for 35\% of all industrial assembly tasks \cite{A19}, representing a canonical case in manufacturing systems\cite{A20}. We therefore selected the reducer—one of the most representative products in gear-based assembly—as our experimental testbed. The $R^2AMS$  consists of a set of robot workcells, as shown in Figure \ref{fig2}A, which is tasked with assembling a batch of reducer products. To further evaluate the generalization capability of the LIRL policy, we designed two dynamic scenarios: one introduces uncertainty in robot operation time due to noise, and the other involves robot failures that lead to uncertainty in production resource availability. The optimization objective of $R^2AMS$ is formulated as a multi-objective function targeting the minimization of both makespan and robotic energy consumption, which is expressed as a weighted combination of these two criteria. An individual reducer assembly task encompasses five stages: \textit{place bottom shell}, \textit{small gear assembling}, \textit{large gear assembling}, \textit{top bottom shell} and \textit{move reducer to buffer},as shown in Figure \ref{fig2}B.  Each robot workcell is capable of performing a complete reducer assembly task. Consequently, $R^2AMS$ represents a fully flexible production and manufacturing system.

\subsection*{Decision variables}

The decision-making space for $R^2AMS$ cross-domain optimization is a hybrid discrete-continuous space, as shown in Figure \ref{fig2}C. The hybrid action $a = (a_d, a_c)$ contains a discrete tuple $a_d = \langle p_{ij}, q_k \rangle$ that assigns a reducer job to a workcell in every operation and a continuous vector $a_c = u_k^{ij} $ that specifies the spline interpolation parameters of the robot trajectory (Methods). The structure of the solution space $T_i$ represents the $i-th$ reducer, $p_{ij}$ represents the $j-th$ operation of the $i-th$ reducer, and $u_k^{ij}$ represents the trajectory parameters of the robot workcell $q_k$
when completed $p_{ij}$. The logical specification encodes (i) capacity constraints, (ii) precedence constraints across the five operations, and (iii) joint-velocity-limit (Kinematics) constraints for trajectories.

The execution trajectory of a robot workcell significantly influences both operational processing time and energy consumption. Simulation experiments quantified the energy consumption profile of a single robot workcell across operations (Figure \ref{fig2}D). The results demonstrate a non-linear relationship between energy consumption and task completion time for a specific operation: At shorter durations, high-speed motion with frequent acceleration/deceleration elevates energy consumption; as completion time increases, reduced acceleration/deceleration minimizes non-functional losses through smoother joint trajectories; however, at excessively prolonged durations, sustained posture maintenance increases non-functional energy consumption, leading to renewed energy escalation. The nonlinear energy-time relationship in robotic operations underscores the need for cross-domain optimization of CPS.

\subsection*{Optimization objectives}

We minimize a convex combination $R = -(\alpha \times C_{makespan} + (1 - \alpha)\times E_{total})$ where $\alpha \in \{0.1, 0.2, 0.3, 0.4, 0.5, 0.6, 0.7, 0.8, 0.9\}$. Makespan is measured in seconds and energy in Joule, both normalized to zero mean and unit variance before weighting.

\subsection*{Baselines}

We compare LIRL with seven baselines representative of current practices:

\begin{itemize} 
    \item $Energy\text{--}opt$  and $Time\text{--}opt$: Hierarchical schedulers that first fix the robot velocities to energy-optimal or time-optimal values and then solve mixed-integer linear programs for task allocation. 
    \item HyAR\cite{A21}: Latent-space reinforcement learning with conditional variational autoencoder (CVAE) pre-training. 
    \item PDQN and HPPO: Specialized reinforcement learning algorithms for hybrid-action  spaces. 
    \item SAC-Lag\cite{A24}: Soft Actor-Critic with Lagrangian constraint relaxation. 
    \item CPO\cite{A25}: Constrained policy optimization enforcing safety constraints directly in policy space via second-order approximation. 
\end{itemize}

All RL methods are trained for 1000 environment steps and are averaged over 10 random seeds.
\begin{figure}[!ht]
\centering
\includegraphics[width=\linewidth]{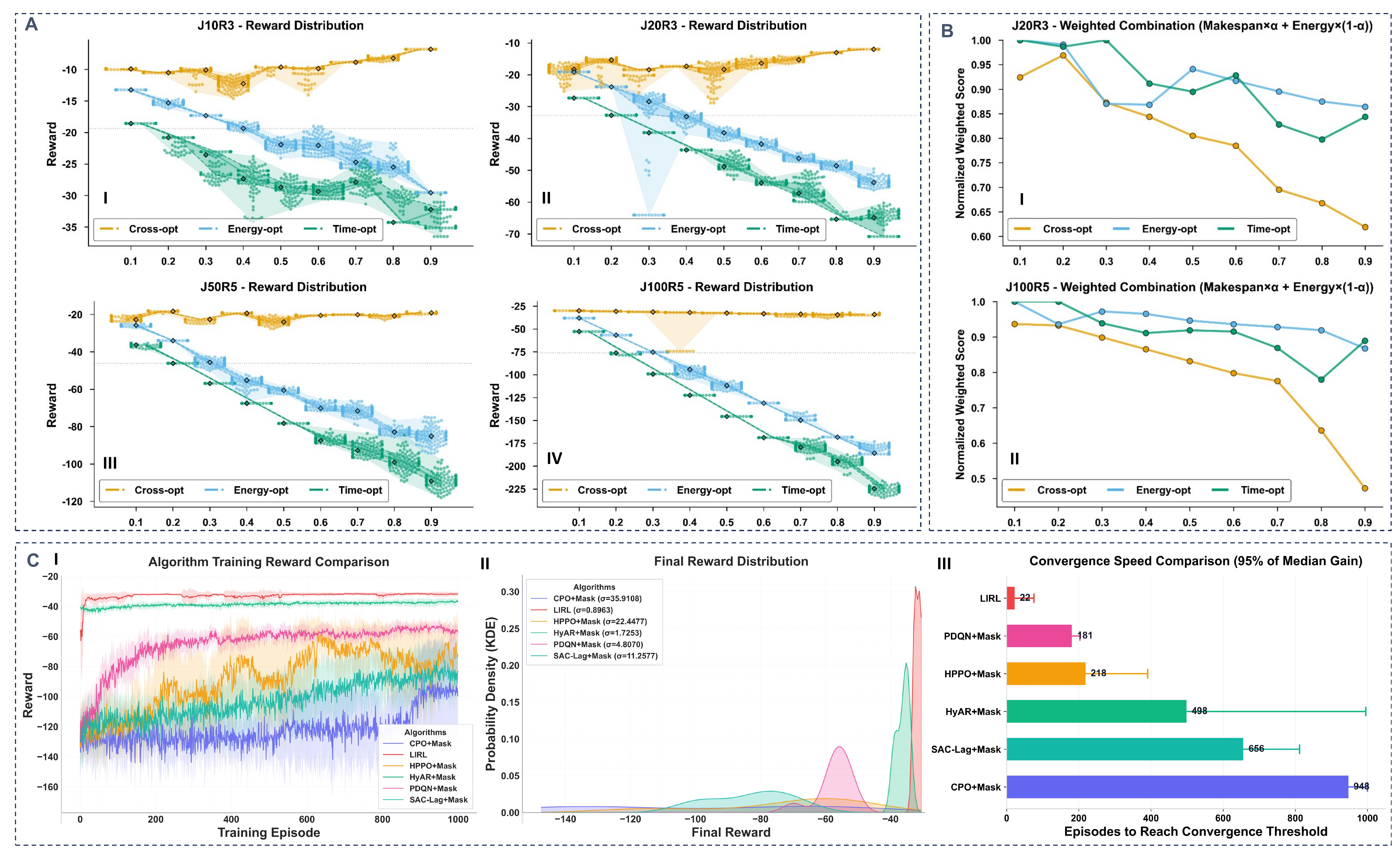}
\caption{Results. \textbf{A}. The comparison of cross-domain optimization(cross-opt) solved by LIRL with Energy-opt and Time-opt methods. Panels I-IV show the reward distribution for varying weights across four scales, with a higher reward denoting superior optimization performance. \textbf{B}. Performance comparison of different optimization methods, measured by a weighted sum of normalized makespan and energy consumption. \textbf{C.I}. It presents a comparison of the training curves between LIRL and the baselines. \textbf{C.II}. The standard deviation of the post-convergence reward distribution quantifies its concentration, with smaller values indicating more stable convergence. \textbf{C.III}. Comparison of convergence performance among different algorithms. The numbers on each algorithm indicate the training episodes required for convergence, defined as the point after which the value remains stable for 95\% of the final value. A lower number corresponds to faster convergence.  }
\label{fig3}
\end{figure}
\subsection*{Simulation Results: overall performance}

The final performance across four production scales, namely $J10R3$, $J20R3$, $J50R5$, and $J100R5$, with different weightings is illustrated in Figure \ref{fig3}A.  Here, $J100R5$ represents 100 reducer assembly tasks carried out in 5 robotic workcells.

The empirical results reported in Figure \ref{fig3}A and Figure \ref{fig3}B can be synthesized into five principal findings, which collectively corroborate the effectiveness and scalability of the proposed $Cross\text{--}opt$ (LIRL) strategy for multi-scale flexible assembly systems.

    Across all four production scales ($J10R3$, $J20R3$, $J50R5$, and $J100R5$) and for the entire weighting range $\alpha\in[0.1,0.9]$, $Cross\text{--}opt$ consistently attains the highest cumulative reward.  
	Its confidence intervals fully envelope those of $Energy\text{--}opt$ and $Time\text{--}opt$, and the linear-regression slopes of its reward curves are the smallest.  
	These observations indicate that Cross-opt persistently approaches the global optimum of the feasible space for the bi-criteria objective (makespan versus energy), whereas the baselines converge to sub-optimal solutions.
	
	The rewards of $Energy\text{--}opt$ and $Time\text{--}opt$ decrease monotonically with increasing~$\alpha$, and the decline steepens markedly for larger problem instances.  
	In sharp contrast, $Cross\text{--}opt$ exhibits an almost flat reward curve.  
	For the $J20R3$ instance, when $\alpha$ is varied from $0.1$ to $0.9$, $Cross\text{--}opt$ fluctuates by merely $4.67\%$, whereas $Energy\text{--}opt$ and $Time\text{--}opt$ deteriorate by $54.21\%$ and $62.38\%$, respectively.  
	Hence, $Cross\text{--}opt$ can operate stably in workshops where the preference weights are uncertain or dynamically adjusted.

	After normalizing the composite objective 
	$\mathcal{L}=\alpha\,C_{\text{makespan}}+(1-\alpha)\,E_{\text{total}}$, 
	we observe the following:
	On the medium-scale $J20R3$ case, $Cross\text{--}opt$ achieved performance improvements of  $14.89\%$ and $10.11\%$ over $Energy\text{--}opt$ and $Time\text{--}opt$, respectively, at $\alpha=0.5$.  
	The advantage widens to $44.33\%$ and $36.47\%$ at $\alpha=0.9$. On the large-scale $J100R5$ case, the same tendency persists and does not attenuate with the increased number of tasks and robots. These evidences confirm that Cross-opt can be transferred across scales with (approximately) linear growth in computational effort.

	The baseline methods search solely in the \emph{cyber layer} ( job-station assignment and sequencing), whereas the \emph{physical layer} ( robot trajectory and dynamics) remains fixed.  
	Consequently, the degrees of freedom are insufficient when $\alpha$ changes, leading to a rapid degradation of their Pareto frontiers.  
	$Cross\text{--}opt$, empowered by reinforcement learning, jointly explores both layers, thereby performing high-dimensional, co\-operative search that continuously reshapes the Pareto frontier and preserves the performance upper bound.
	
    For manufacturing scenarios with uncertain or frequently re-tuned preference weights, \textsc{Cross-opt} can substantially reduce model maintenance and parameter calibration costs. As the scale of flexible manufacturing expands, traditional single-layer or serial optimization frameworks become susceptible to the \emph{curse of dimensionality}, whereas the LIRL paradigm offers a viable, scalable alternative.

In summary, $Cross\text{--}opt$ outperforms $Energy\text{--}opt$ and $Time\text{--}opt$ in terms of optimality, robustness, and scalability, thereby substantiating the potential of cross-domain optimization for complex manufacturing systems.

\subsection*{Convergence and sample efficiency}

Because every production-scheduling constraint is hard by definition, designing a well-behaved reward function is far from trivial. In practice, the original baseline agents could not eliminate constraint violations and therefore failed to converge. To guarantee feasibility at every decision step, each baseline was augmented with an invalid action mask\cite{A26} layer, yielding the variant referred to as baseline+Mask,such as HyAR+Mask in  Fig. \ref{fig3}C. These masked baselines were then benchmarked against the proposed LIRL, under $J100R5$ with $\alpha=0.5$.

The quantitative results reported in Figure \ref{fig3}C.I, Figure \ref{fig3}C.II, and Figure \ref{fig3}C.III give rise to five main observations.

	Throughout training LIRL attains the largest average return among all competitors and culminates in the highest final reward (Figure \ref{fig3}C.I). Given that all algorithms are subject to identical safety constraints, these results indicate that the policy returned by LIRL is closest to the global optimum
	within the feasible action space.
	
	Kernel Density Estimation of the terminal returns (Figure \ref{fig3}C.II) yields a standard deviation of $\sigma = 0.8963$ for LIRL, which is the smallest among all baselines. The tight concentration across 10 random seeds implies superior run-to-run consistency and, by extension, higher robustness in real-world deployments.
	
	As shown in Figure \ref{fig3}C.III, LIRL requires only 22 episodes to reach $95\%$ of its final performance,	whereas the next fastest method, PDQN+Mask, needs 181 episodes. SAC-Lag+Mask and CPO+Mask further lag behind at 656 and 948 episodes, respectively.
	Hence, LIRL accelerates convergence by an order of magnitude (10–40$\times$), primarily because exploration is confined to the constraint-satisfying sub-space, avoiding low-reward actions that violate safety.

	HyAR+Mask exhibits a smooth early climb but relies on a CVAE pre-training stage that already exploits LIRL trajectories, yielding a near-optimal starting point; it is nevertheless surpassed by LIRL in later stages. Penalty-based methods (SAC-Lag+Mask, CPO+Mask) must repeatedly sample infeasible
	actions and dynamically adjust Lagrange multipliers, leading to slow and oscillatory learning curves.  Although PDQN and HPPO profit from masking, they still depend on $\varepsilon$-greedy or Gaussian exploration, resulting in larger terminal variance ($\sigma \approx 4$–$22$).

	Explicitly encoding safety as an action-generation prior while retaining a reward-maximization objective yields a favorable triple outcome: high efficiency, high return, and high stability. In domains where interaction is costly or failure is unacceptable (e.g.\ robotic manipulation or autonomous driving),
	the order-of-magnitude speed-up achieved by LIRL translates directly into reduced tuning costs and enhanced operational reliability.

In summary, LIRL simultaneously dominates strong baselines on optimality, stability, and convergence speed, thus furnishing a promising solution for high-dimensional, safety-constrained control.




\subsection*{Ablation study}

This study employs the deep deterministic policy gradient\cite{A27} (DDPG) framework to compare two action-processing strategies under an identical agent policy. The first strategy applies a projection method grounded in LIRL, whereas the second adopts an invalid-action masking mechanism. Experiments are conducted on four task configurations—$J10R3$, $J20R3$, $J50R5$, and $J100R5$—with the weight parameter $\alpha$ fixed at 0.5.

\begin{figure}[ht]
\centering
\includegraphics[width=\linewidth]{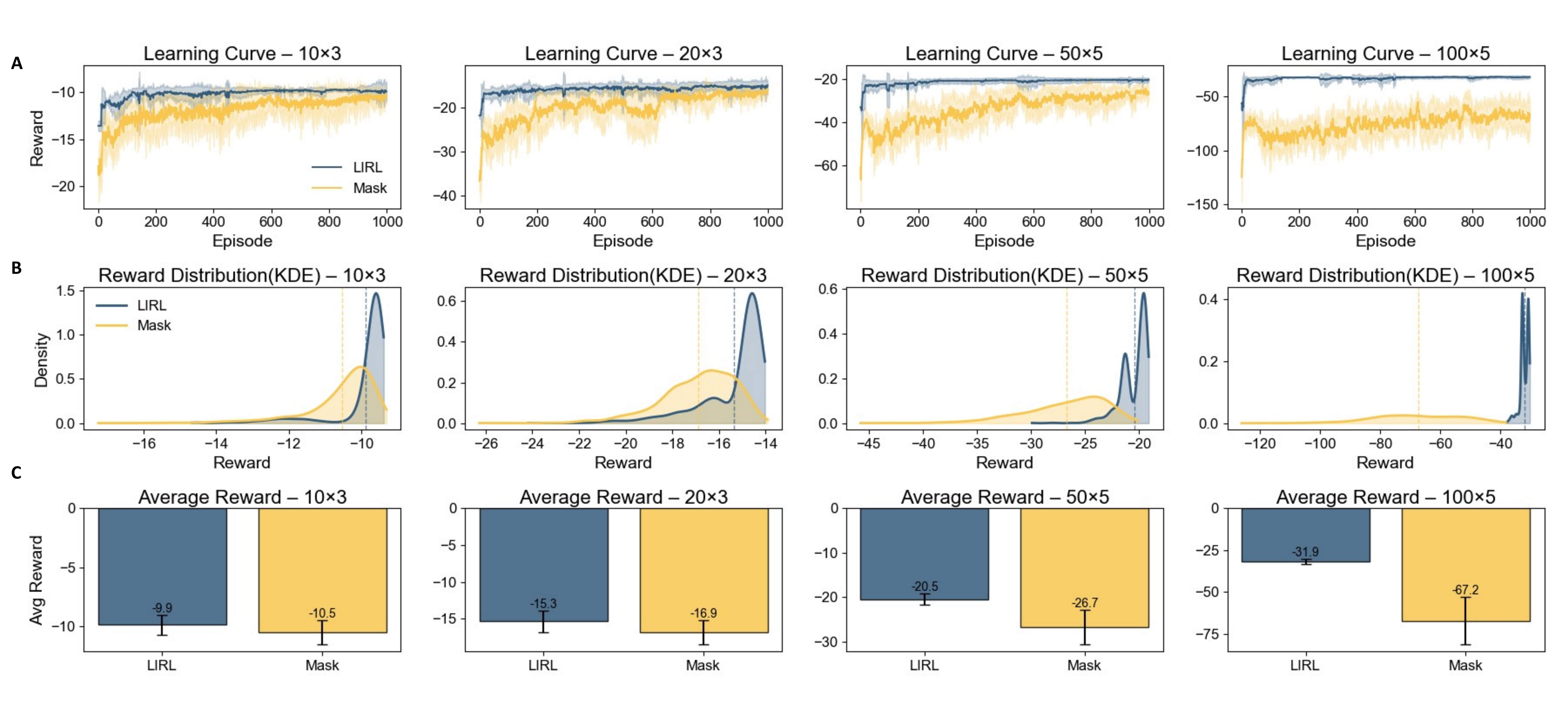}
\caption{Ablation results. \textbf{A}. The learning curve comparison of four scales. \textbf{B}. The post-convergence reward distribution. \textbf{C}. Comparison of the average rewards of LIRL and Mask under different scales }
\label{fig4}
\end{figure}


The learning curves in Figure \ref{fig4}A reveal a markedly steeper ascent for LIRL during the first 150~episodes, after which the return quickly plateaus.  Conversely, the Mask baseline exhibits a step-wise, slow climb and has not fully stabilized by the end of the 1\,000-episode horizon.  
The gap is particularly salient in the high-dimensional tasks (50\,$\times$\,5 and 100\,$\times$\,5), underscoring the superior sample efficiency of LIRL.

The final-episode average returns (Figure \ref{fig4}C) demonstrate a systematic advantage of LIRL that widens monotonically with task scale: 10\,$\times$\,3: $-9.9$ vs.\ $-10.5$; 20\,$\times$\,3: $-15.3$ vs.\ $-16.9$; 50\,$\times$\,5: $-20.5$ vs.\ $-26.7$; 
100\,$\times$\,5: $-31.9$ vs.\ $-67.2$. The reward gap therefore expands from $0.6$ to $35.3$, indicating that LIRL scales more gracefully to exponentially larger state–action spaces.

Kernel density estimates of the return distribution (Figure \ref{fig4}B) show that LIRL yields a higher and sharper peak, while its tails are substantially curtailed; the empirical standard deviation is roughly 50\% that of the Mask baseline.  
Hence, LIRL not only improves the expected performance but also reduces variance, evidencing greater training stability.

The projection confines policy outputs to the convex set of legal actions, thereby eliminating the zero-gradient regions introduced by masking and retaining continuous, informative gradients—accounting for the accelerated convergence.
By suppressing exploration of infeasible actions at the source, LIRL increases the density of informative samples, translating into higher final returns under identical interaction budgets. As the action space grows, Mask still performs forward passes over all actions before zeroing invalid ones, incurring redundant computation and additional noise.  LIRL mitigates both issues, yielding super-linear relative gains in large-scale scenarios.

Across convergence speed, asymptotic return and stability, LIRL delivers statistically significant improvements over Invalid-Action Masking, with benefits that amplify as task complexity increases.  
These findings substantiate the suitability of LIRL for large-scale constrained reinforcement learning problems.

\begin{figure}[!t]
	\centering
	\includegraphics[width=\linewidth]{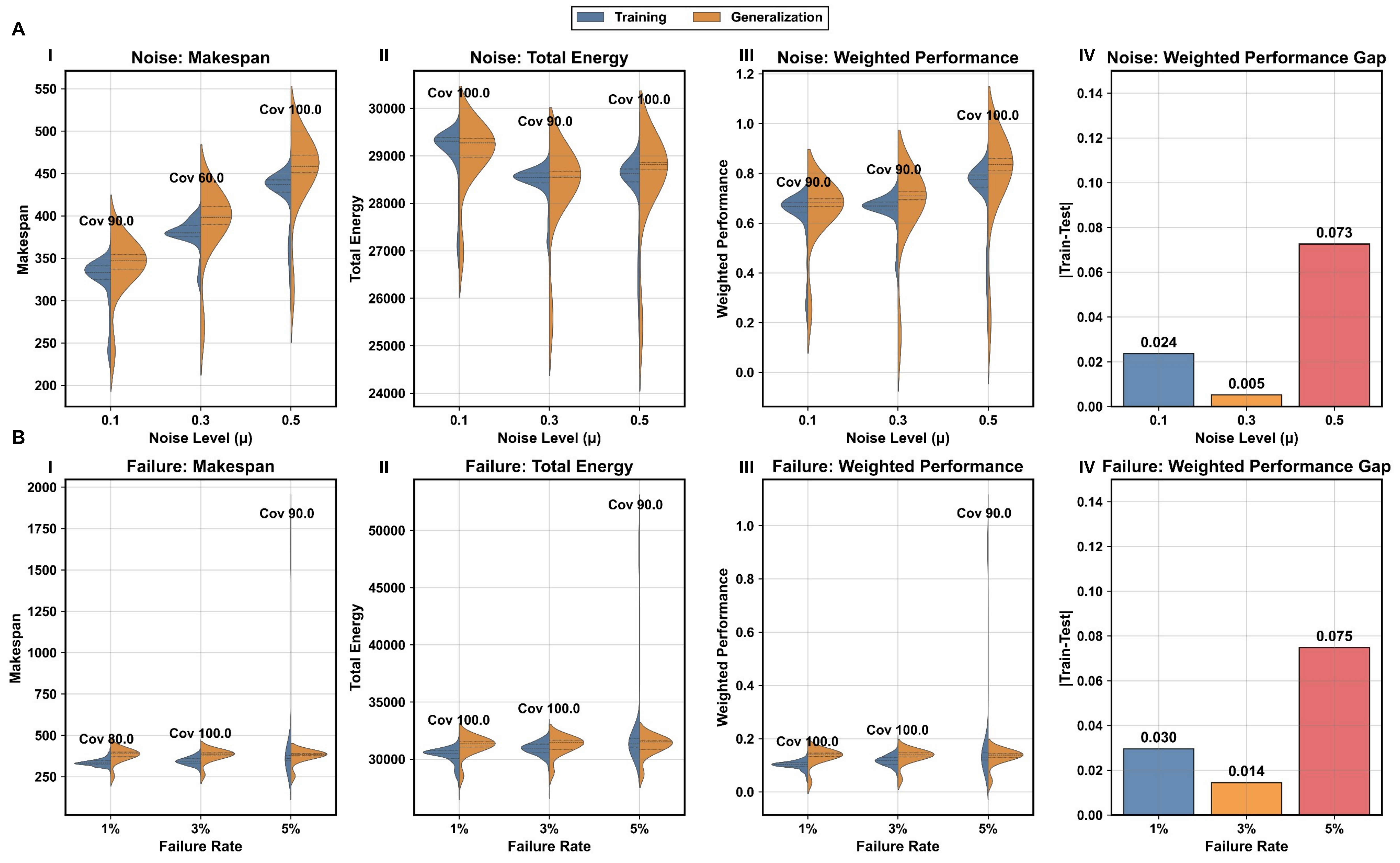}
	\caption{Results of the robustness testing experiment. The "Training" performance refers to models trained and evaluated at specific perturbation levels, whereas the "Generalization" performance is assessed on a single model trained under disturbance-free conditions. "Cov" indicates that the solution obtained through "Generalization" covers the range of the "Training" solution.  \textbf{A}. Performance of LIRL-trained strategies under different noise levels.  \textbf{B}. Performance of LIRL-trained strategies under different robot failure rate. }
	\label{fig5}
\end{figure}

\subsection*{Robustness to stochastic disturbances}

Real-world manufacturing lines suffer from uncertain processing times and unexpected machine faults. We therefore introduce stochastic noise perturbation and exogenous failure into simulation.
\begin{itemize}
    \item Duration time uncertainty. The processing time of each operation is perturbed by zero-mean Gaussian noise with standard deviations set to 0.1$\mu$, 0.3$\mu$ and 0.5$\mu$, where $\mu$ represents the average completion time of each operation, $\mu =t_{min}+0.5\times(t_{max}-t_{min})$, $t_{min}$, $t_{max}$ denote the minimum and maximum theoretical duration times of each operation, respectively.
    \item  Robot breakdown. Robots are assumed to break down with probabilities of 1 \%, 3 \%, and 5 \% per robot workcell. Each failure triggers a repair period lasting between 1.5 × and 3 × the nominal processing time of the interrupted task. At most ten failures are allowed within a single training episode.
\end{itemize}

Figure \ref{fig5} shows the complete performance distributions obtained with the proposed LIRL policy under two typical out--of--distribution (OOD) perturbations: additive input noise and stochastic machine failures.  The curves marked \textit{Training} denote in–distribution performance, whereas \textit{Generalization} reports the same policy evaluated on perturbed test instances.   Across noise levels $\mu = 0.1\!-\!0.5$ and failure rates $1\%\!-\!5\%$, the \textit{Generalization} curves cover more than $90\%$ of the \textit{Training} range; for $\mu = 0.5$ and a $3\%$ failure rate the coverage even reaches $100\%$.  For all three metrics—\textit{Makespan}, \textit{Total Energy} and \textit{Weighted Performance}—the discrepancy between the two means never exceeds $0.075$, implying that, under medium-to-high OOD perturbations, the policy remains well confined to the training distribution with only moderate variance.

\begin{figure}[H]
	\centering
	\includegraphics[width=\linewidth]{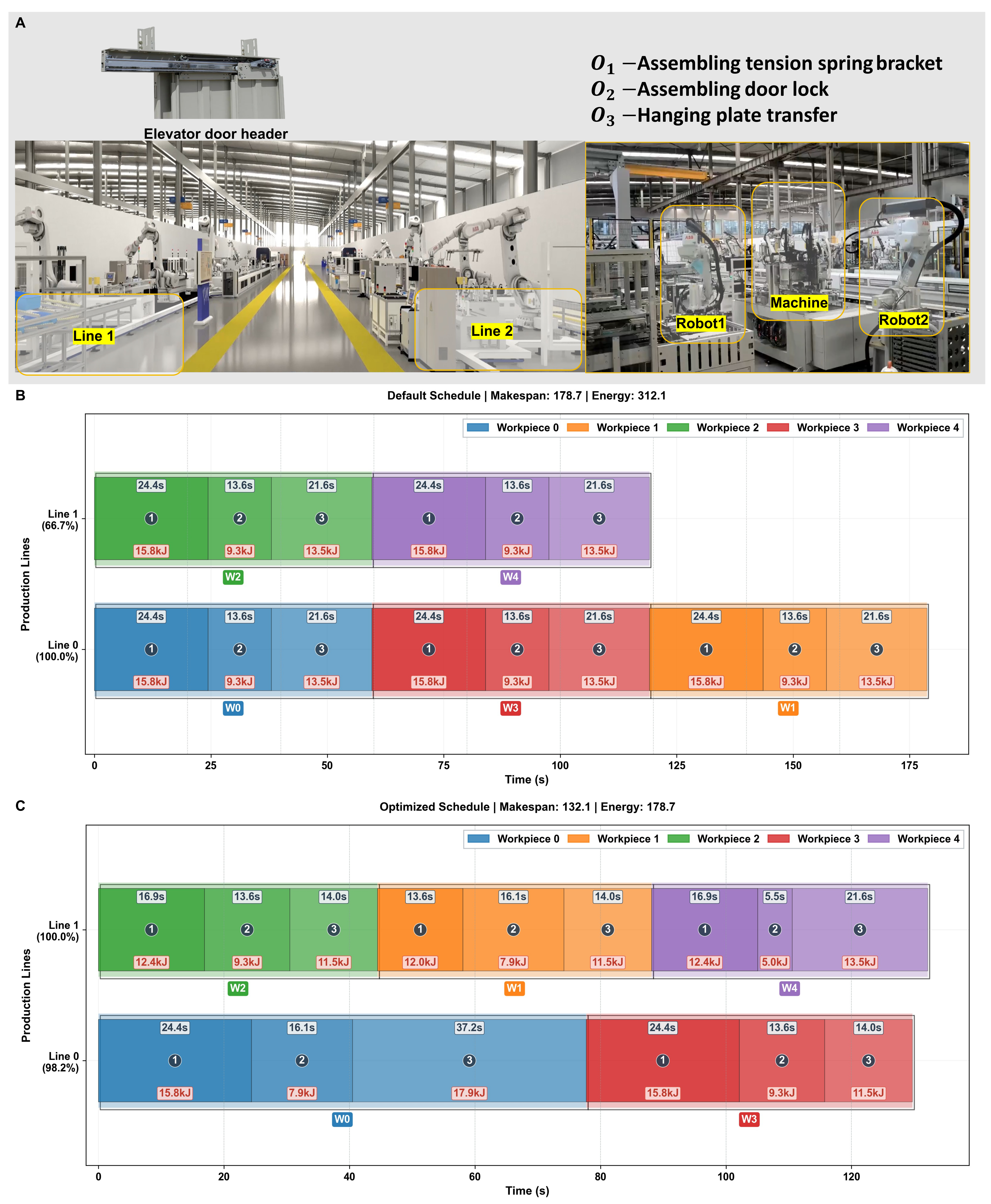}
	\caption{Overview and result of validation in an actual elevator door head manufacturing factory.  \textbf{A}. Overview of the elevator door-header plant. The factory contains two parallel assembly lines, and each line is equipped with an identical hanging-plate assembly workcell. Every workcell comprises two industrial robots and one dedicated assembly machine.  The robots collaborate to execute three sequential operations for each workpiece:  (O$_1$) tension-spring-bracket assembly,  (O$_2$) door-lock assembly, and  (O$_3$) hanging-plate transfer. \textbf{B}. The original scheduling plan of the factory.  \textbf{C}. The optimized scheduling plan generated by LIRL.}
	\label{fig6}
\end{figure}

As perturbations intensify, both \textit{Training} and \textit{Generalization} distributions shift rightward simultaneously, while the pairwise distance between them remains nearly constant.  In particular, the energy metric attains $100\%$ coverage at all three failure levels, demonstrating that the learned policy can stabilize energy consumption without compromising production tempo.

When $\mu = 0.5$ or the failure rate rises to $5\%$, the mean gap increases once to about $0.07$–$0.08$; nevertheless, the coverage is still no less than $90\%$.  Hence the policy does not collapse but rather enters an acceptable-degradation zone. A single LIRL policy trained on clean data can be deployed directly over a wide spectrum of perturbation magnitudes, eliminating the need for multiple models and repeated hyper-parameter tuning in industrial settings.  Even under sensor drift, equipment ageing, or other unforeseen factors, the policy sustains near-optimal cycle time and energy usage, thereby enhancing line availability and overall efficiency.

In summary, the empirical results corroborate previous evidence: LIRL attains an average coverage of $92.78\%$ with a mean performance loss below $7.5\%$ across random noise and machine-failure scenarios, demonstrating strong OOD generalization and industrial-grade robustness.



\subsection*{Field study: real-world deployment in an elevator door-header factory}

The LIRL scheduler was deployed in a commercial elevator door--header factory (Figure \ref{fig6}A) to \textit{(i)} assign workpieces to two parallel production lines and  \textit{(ii)} orchestrate the three elemental robot operations in the hanging-plate assembly workcell.  Each workpiece requires a fixed process plan with three sequential operations: $\mathrm{O}_{1}$ (tension-spring-bracket assembly),
$\mathrm{O}_{2}$ (door-lock assembly), and
$\mathrm{O}_{3}$ (hanging-plate transfer). The policy was trained entirely offline using historical execution logs; no online exploration was conducted during deployment to ensure zero disruption to shop-floor operations A detailed description of the hardware architecture is given in the Supplementary Materials.

In a pilot deployment, we evaluated an order comprising five workpieces. Under identical conditions of equipment, matched production environment, and personnel allocation, scheduling plans were generated and executed by both the existing rule-based scheduling system (serving as the baseline) and the proposed LIRL strategy. To support result re-producibility and facilitate extended statistical analysis, monitoring devices were deployed on the production line to systematically record the following data: (i) robot joint angles and power consumption under high-frequency sampling; (ii) the time required for the robot to complete each operation; and (iii) all constraint verification processes and associated interlock statuses. Primary outcomes are (i) makespan (s), (ii) aggregate electrical energy (kJ), and (iii) line utilization. 

Schedules produced by the plant’s rule-based dispatcher and by the proposed LIRL policy are executed and compared (Figure \ref{fig6}B and \ref{fig6}C, respectively). The LIRL schedule completes the order in $132.1\,\mathrm{s}$, a $26.1\%$ reduction relative to the $178.7\,\mathrm{s}$ baseline. Aggregate electrical energy drawn by the two robots drops from $312.1\,\mathrm{kJ}$ to $178.7\,\mathrm{kJ}$, yielding a $42.7\%$ saving. The rule-based logic keeps Line~0 saturated while leaving Line~1 idle for roughly one–third of the horizon (utilization $66.7\%$).	LIRL raises utilization to $98.2\%$ (Line~0) and $100\%$ (Line~1), achieving a nearly balanced flow.

The above improvements originate from the multi-objective reward adopted during offline training, which jointly makespan, cumulative energy, and workload imbalance. The learned policy not only shortens production time but also smooths robot velocity profiles, thereby lowering energy demand without compromising throughput. Policy inference takes merely a few seconds, rendering the approach suitable for rolling-horizon rescheduling in discrete-part manufacturing.

The field trial demonstrates that an offline-trained LIRL scheduler can simultaneously improve makespan, energy efficiency, and line balance in a multi-line, multi-robot environment. These results substantiate the viability of reinforcement-learning-driven production control as a plug-and-play component for digital, low-carbon factories.

\subsection*{Simulation-based evaluations in other domains: Transportation and Micro-grids}
\begin{figure}[!ht]
	\centering
	\includegraphics[width=\linewidth]{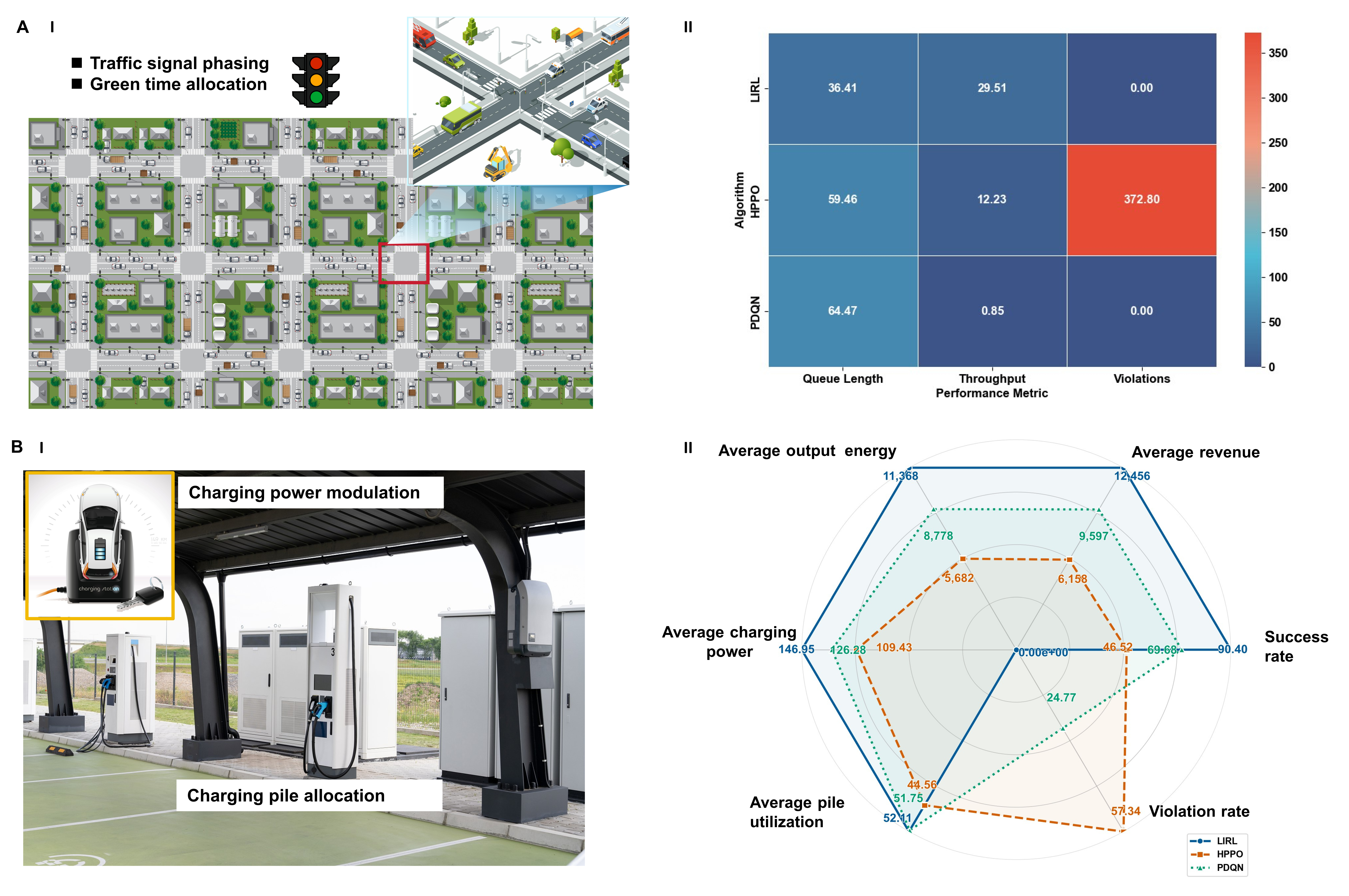}
	\caption{Results of the extended experiment. \textbf{A.I}. The urban road network with a 3×5 grid layout. \textbf{A.II} presents the performance comparison of the LIRL, PDQN, and HPPO algorithms using these metrics: Queue Length, average number of vehicles waiting at each intersection, sampled every 10 seconds; Throughput, average number of vehicles passing through each intersection, sampled every 10 seconds; Violations, total number of traffic rule violations across all intersections during the one-hour simulation. \textbf{B.I}. Overview of EV-charging station. \textbf{B.II} presents the performance comparison of the LIRL, PDQN, and HPPO algorithms using these metrics: Average Revenue, daily income, excluding costs from charger depreciation; Average Output Energy, total energy output per day; Average Charging Power, mean charging power over the day. Average Station Utilization, average utilization rate of the charging stations; Violation Rate, degree to which the scheduling decisions violate operational constraints. Success Rate, probability that an arriving vehicle successfully starts a charging session. }
	\label{fig7}
\end{figure}

The applicability of LIRL was examined in two representative CPS domains: transportation and micro-grids. Full scenario specifications are provided in the Supplementary Material.

Based on the quantitative results shown in Figure \ref{fig7}, we report the following findings.  
For clarity, the discussion is organized according to two canonical CPS tasks, i.e.\ urban traffic control and EV–charging micro-grids, followed by a cross-scenario synthesis.

\noindent\textbf{Urban traffic control}. Compared with the baselines, LIRL reduces the \emph{network‐wide average queue length} by $43.5\,\%$ against PDQN and by $38.8\,\%$ against HPPO. Meanwhile, the \emph{system throughput} is boosted to
$29.51\times10^{2}\,\text{veh}/\text{h}$, i.e.\ $34.7\times$ PDQN and $2.41\times$ HPPO (Figure \ref{fig7}A.II). Throughout the entire evaluation horizon, LIRL introduces \emph{zero} signal‐phase conflicts and no green‐time violations, whereas HPPO records $372.8$ violations.  
PDQN is violation‐free but with low efficiency.  The results demonstrate that LIRL simultaneously mitigates congestion and maintains the required safety envelope, hence achieving an efficiency–safety dual optimum.

\noindent\textbf{EV charging micro-grid}

LIRL attains an average daily revenue of \mbox{12\,456}, outperforming PDQN and HPPO by $29.8\,\%$ and $102\,\%$, respectively(Figure \ref{fig7}B.II). The \emph{charger utilization ratio} and the \emph{average output power} rise to $52.71\,\%$ and $146.95\,$kW; both metrics exceed PDQN and HPPO, confirming LIRL’s superior resource allocation under fluctuating loads. The \emph{per‐vehicle success rate} reaches $90.40\,\%$ with zero violations of transformer capacity or current limits, in contrast to PDQN ($24.77\,\%$) and HPPO ($57.34\,\%$) violation rates.  
Hence, LIRL yields the best trade-off among profit, utilization, and operational safety.

\noindent\textbf{Cross-scenario synthesis}

By means of adaptive dual variables, LIRL balances optimization and feasibility, overcoming the classic efficiency–safety dilemma in RL. The zero‐violation logs indicate that LIRL preserves constraint adherence during both learning and evaluation phases, a prerequisite for real-world deployment on critical infrastructures. Consistent superiority across heterogeneous tasks suggests that the LIRL framework can transfer to other energy and transport domains, positioning it as a promising candidate for large-scale CPS optimization.

Experimental evidence confirms that LIRL  
(i) maximizes system efficiency while \emph{strictly} meeting operational constraints;  
(ii) generalizes across tasks of different scales and structures; and  
(iii) offers a compelling empirical basis for \emph{safe-to-deploy reinforcement learning}.

\section*{Discussion}

This work advances the state of cyber–physical cross-domain optimization by introducing logic-informed reinforcement learning, a generic framework that fuses declarative constraint reasoning with gradient-based policy learning. Through a projection operator that maps latent actions onto the dynamically constructed feasible manifold, LIRL converts  standard policy-gradient algorithm into a constraint-preserving optimizer for hybrid  discrete-continuous action spaces. Extensive experiments on a representative robotic assembly line demonstrate that LIRL (i) improves makespan–energy trade-offs by 36.47–44.33\% at most over hierarchical schedulers, (ii) surpasses recent constrained and hybrid-action RL baselines while eliminating all training-time violations, (iii) accelerates convergence by a factor of ten to forty, and (iv) maintains robustness under stochastic task durations and machine faults. The findings have several implications and limitations that merit discussion.

The projection mechanism decouples feasibility from reward shaping, thereby removing a major practical impediment to deploying RL in safety-critical CPS. Plant operators no longer need to hand-craft intricate penalty functions or retune them whenever the environment changes; instead, they simply encode domain knowledge as first-order logic templates. Because these templates are declarative, the same RL backbone can be ported across domains with minimal engineering effort. Moreover, LIRL accelerates learning by avoiding invalid exploration. This property reduces the amount of simulated or real-world interaction data required to reach competent performance, shortening the commissioning time of RL-enabled industrial controllers. In the context of robotics, the guaranteed feasibility also eliminates safety cages or hardware interlocks during policy refinement, lowering deployment costs.

Despite these strengths, several limitations warrant attention. Our benchmark test used 100 products and 5 work units—a scale that already challenges traditional optimization methods but remains moderate compared to industrial settings involving hundreds of workstations. Current physical constraints are limited to first-order linear forms; incorporating more complex constraints (e.g., multi-robot collision avoidance) could incur substantial computational cost, possibly requiring incremental or sampling-based methods for real-time decision-making.
Although empirically and locally we show that the projection operator preserves policy-gradient fixed points, a general convergence proof for non-convex hybrid manifolds is still open. Extending projected stochastic optimization to hybrid action spaces is a promising direction. The framework also assumes stationary constraints, yet real-world CPS often face changing regulations or hardware updates. Future work should explore online logic synthesis and lifelong constraint learning for adaptive compliance.

\section*{Methods}

This section details the problem formalization, the logic-informed reinforcement-learning framework, the projection operator, and the experimental protocol. All implementation scripts, configuration files and raw data are released at \url{https://github.com/wanguangxi/LIRL-CPS} to ensure full reproducibility. 
For more implementation details, please refer to the supplementary materials.

\subsection*{Problem formulation}
We cast CPS cross-domain optimization as a constrained Markov decision process (CMDP)
$$
\mathcal{M} \triangleq \langle \mathcal{S}, \mathcal{A}, \mathcal{T}, r, \mathcal{C}, \gamma, \Phi(s,a), \mathbb{F}(s),  \mathbf{z}, \Pi_s \rangle,
$$
where state $s \in \mathcal{S}$ captures the joint cyber–physical status of the plant, action $a \in \mathcal{A}$ is hybrid, transition kernel $\mathcal{T}$ encodes stochastic process dynamics, $r(s,a)$ is the primary scalar reward, $\gamma \in (0,1)$ the discount factor, and $\mathcal{C} \triangleq \{c_k(s,a) \leq 0\}_{k=1}^K$ denotes $K$ hard constraints. $\mathcal{A} = \mathcal{A}^d \times \mathcal{A}^c$ is the Cartesian product of a finite discrete set $\mathcal{A}^d$ and a continuous box $\mathcal{A}^c \subset \mathbb{R}^P$. $\Phi(s,a) := \bigwedge_{k=1}^{K} \phi_k(s,a)$ – conjunctive set of $K$ state-dependent constraints. Each $\phi_k$ is assumed to be twice continuously differentiable in its continuous arguments and can be evaluated in $O(1)$ time for the discrete part. $\mathbb{F}(s) := \{a \in \mathcal{A} \mid \Phi(s,a)=\text{True}\}$ – state-conditioned feasible set. By engineering design $\mathbb{F}(s) \neq \emptyset$ for every $s$. $\mathbf{z} \in \mathbb{R}^L$ – latent “united” action produced by the policy network; ${L \ll |\mathcal{A}|}$. $\Pi_s:\mathbb{R}^L \rightarrow \mathbb{F}(s)$ – state-conditioned projection operator; $a = \Pi_s(z)$. $\pi_\theta(z|s)$ – latent-space stochastic policy with parameters $\theta$; the executed policy is $\pi^\Pi_\theta(a|s)=\int \mathbb{I}\{\Pi_s(z)=a\} \pi_\theta(z|s) \, dz$. Our objective is to maximize the discounted return $J(\pi^\Pi_\theta) = \mathbb{E}_\pi[\sum \gamma^t r_t]$ subject to $c_k(s_t,a_t) \leq 0$ $\forall k,t$.

\subsection*{Projection}

For any state $s$ and latent vector $z$, $\Pi_s$ is computed in two independent steps:

Discrete action projection. Let $u = f_d(z) \in \mathbb{R}^{|\mathcal{A}^d|}$ be logits decoded from $z$. We solve

\begin{equation}\label{eq:projection_dis} 
    a^d = \arg\max_{a \in \mathcal{A}^d} \langle u,a \rangle \quad \text{s.t.} \quad \phi_\text{cap}(s,a)=\text{True}, \phi_\text{prec}(s,a)=\text{True},
\end{equation}
where $\phi_\text{cap}$, $\phi_\text{prec}$  represent the capacity constraint and the precedence constraint respectively. Because the constraints are linear and totally unimodular, the $\arg\max$ is equivalent to a linear-assignment problem solvable in $O(|\mathcal{A}^d|^3)$ by the Hungarian algorithm.

Continuous action projection. Let $v = f_c(z) \in \mathbb{R}^P$ be preliminary spline parameters. We solve a strictly convex quadratic program

\begin{equation}\label{eq:projection_con} 
a^c = \arg\min_{x \in \mathbb{R}^P} \|x - v\|^2 \quad \text{s.t.} \quad \phi_\text{kin}(s,a^d,x) \leq 0, \phi_\text{col}(s,a^d,x) \leq 0,
\end{equation}
where $\phi_\text{kin}$, $\phi_\text{col}$ represent the kinematic constraints and collision constraints of the robot respectively. Slater’s condition holds by design, hence the KKT system has a unique solution; we employ operator splitting quadratic program\cite{A28}, OSQP with average complexity $O(P^2)$.

Finally $\Pi_s(z) := (a^d,a^c)$. The detailed algorithms can be found in the Supplementary Material.

\subsection*{Training}

Each episode runs for at most 500~steps. When the replay buffer holds more than 500 transitions, the agent updates both actor and critic for 20~gradient steps per environment
step.  Models are check pointed every 10~episodes for later evaluation. Unless otherwise specified, all reported metrics are averaged over $10$~random seeds to ensure statistical significance.
The LIRL policy using a single NVIDIA RTX4090 GPU. The training process takes 0.5 to 2.8 hours depending on problem size.

\subsection*{Constraint satisfaction guarantee}
\textbf{Lemma 1 (Feasibility)} For every state $s$ and latent $\mathbf{z}$, the action $a = \Pi_s(\mathbf{z})$ belongs to $\mathbb{F}(s)$.

\textit{Proof} The discrete step enforces all purely cyber constraints; the continuous QP is solved over the intersection of physical constraints. By construction each $\phi_k$ is satisfied, hence $\Phi(s,a)=\text{True}$ and $a \in \mathbb{F}(s)$. \hfill $\blacksquare$

\subsection*{Regularity of $\Pi$}

\textbf{Assumption 1 (Regular constraints)}
\begin{enumerate}
	\item[(a)] Each continuous constraint $\phi_k(s,a^c)$ is $C^2$ and its gradient with respect to $a^c$ is $L_k$-Lipschitz.
	\item[(b)] For every $s$ the Jacobian of active constraints has full row rank at the optimum of the QP.
\end{enumerate}

\textbf{Proposition 1 (Differentiability)} Under Assumption 1, the mapping $a^c = \Pi_c(s,z)$ is $C^1$ almost everywhere in $z$; its Jacobian is given by implicit differentiation of the KKT conditions. The overall $\Pi_s$ is therefore piecewise-smooth.

\textbf{Corollary 1 (Lipschitz continuity)} There exists $L_\Pi > 0$ such that $\|\Pi_s(z_1) - \Pi_s(z_2)\| \leq L_\Pi \|z_1 - z_2\|$ for all $z_1,z_2$.

\subsection*{Compatibility with policy-gradient methods}

We consider the episodic performance objective
\begin{equation}\label{eq:obj}
	J(\theta) = \mathbb{E}_{s_0, z_{0:T-1}} \left[\sum_{t=0}^{T-1} r(s_t, \Pi_{s_t}(z_t))\right].
\end{equation}

The latent-space likelihood is $\pi_\theta(z_t|s_t)$. Using the standard score-function estimator yields the gradient
\begin{equation}\label{eq:obj_2}
	\nabla_\theta J(\theta) = \mathbb{E}\left[\sum_{t} \nabla_\theta \log \pi_\theta(z_t|s_t) \cdot G_t\right],
\end{equation}
where $G_t$ denotes the cumulative return from $t$. Importantly, $\Pi$ appears only inside the return, not inside the score term, so the estimator remains unbiased provided $\Pi$ is deterministic and measurable.

\textbf{Theorem 1 (Preservation of stationary points)} Assume
\begin{itemize}
	\item $r(s,a)$ is bounded and $C^1$ in its continuous arguments;
	\item $\Pi_s$ satisfies Lemma 1 and Proposition 1;
	\item The latent policy $\pi_\theta$ is $C^1$ and non-degenerate on $\mathbb{R}^L$.
\end{itemize}

Then any $\theta^\star$ that is a stationary point of the projected objective $J(\theta)$ is also a stationary point of the original CMDP objective restricted to feasible actions, i.e., $\nabla_\theta J(\theta^\star)=0 \Rightarrow \theta^\star$ corresponds to a policy whose executed distribution $\pi^\Pi_{\theta^\star}$ cannot be improved within the feasible set by an infinitesimal latent variation.

\textit{Proof} Because $\Pi_s$ is Lipschitz and piecewise-smooth, we can interchange differentiation and expectation (dominated convergence). The chain rule gives
\begin{equation}\label{eq:obj_3}
	\nabla_\theta r(s,\Pi_s(z)) = \partial_a r(s,a)\vert_{a=\Pi_s(z)} \cdot \partial_\theta \Pi_s(z) + 0,
\end{equation}
but $\partial_\theta \Pi_s(z)=\partial_z \Pi_s(z)\cdot\partial_\theta z$, with $\partial_z \Pi_s$ bounded by Proposition 1. Consequently, the projected policy-gradient has the same zero set as the gradient of the constrained objective defined directly over $\mathbb{F}$. \hfill $\blacksquare$

\subsection*{Convergence}

Let $\hat{g}_t$ be the Monte-Carlo estimate of $\nabla_\theta J(\theta)$ at iteration $t$. With a learning rate sequence $\eta_t$ satisfying $\sum \eta_t = \infty$, $\sum \eta_t^2 < \infty$, the update $\theta_{t+1} = \theta_t + \eta_t \hat{g}_t$ constitutes a Robbins–Monro stochastic approximation.

\textbf{Corollary 2 (Almost-sure convergence)} Under Theorem 1 and the usual variance-boundedness condition on $\hat{g}_t$, the parameter sequence $\{\theta_t\}$ converges almost surely to the set of stationary points of $J(\theta)$.

\textbf{Remark 1} If the continuous constraints are further convex and $\Pi_c$ reduces to Euclidean projection, convergence rate matches that of projected stochastic gradient ($O(1/\sqrt{t})$). Non-convex collision constraints break global guarantees, yet empirical results confirm stable convergence in all tested instances.

\bibliography{sample}

\section*{Data and code availability}

All source code of LIRL, logic files and results are on \url{https://github.com/wanguangxi/LIRL-CPS}.

\section*{Acknowledgments}

This work was supported by the National Key Research, and Development Program of China [2024YFB4711103], National Natural Science Foundation of China [U24A20277, 92267205, 62503466, 92467301, 92367301],  Natural Science Foundation of Liaoning Province [2024-MSBA-83, 2025-MS-085], Fundamental Research Project of SIA [2024JC1K10], and the State Key Laboratory of Robotics and Intelligent Systems of China [2025-Z12].

\section*{Author contributions statement}

Guangxi Wan and Peng Zeng conceived the experiments,  Yiyang Liu,  Xiaoting Dong and Hongfei Bai conducted the experiments, Shijie Cui, Qingwei Dong  Dong Li, and Chunhe Song analyzed the results.  All authors reviewed the manuscript.

\end{document}